\documentclass{article}

\usepackage{microtype}
\usepackage{graphicx}
\usepackage{subfigure}
\usepackage{booktabs} 

\usepackage{hyperref}

\usepackage[accepted]{icml2025}

\usepackage{amsmath}
\usepackage{amssymb}
\usepackage{mathtools}
\usepackage{amsthm}

\usepackage[capitalize,noabbrev]{cleveref}

\theoremstyle{plain}

\theoremstyle{definition}

\theoremstyle{remark}

\usepackage[textsize=tiny]{todonotes}

\icmltitlerunning{Confidently Classified Counterfeits}

\begin{document}

\twocolumn[
\icmltitle{Network Inversion for Generating Confidently Classified Counterfeits}



\icmlsetsymbol{equal}{*}

\begin{icmlauthorlist}
\icmlauthor{Pirzada Suhail}{yyy}
\icmlauthor{Pravesh Khaparde}{yyy}
\icmlauthor{Amit Sethi}{yyy}
\end{icmlauthorlist}

\icmlaffiliation{yyy}{Department of Electrical Engineering, IIT Bombay}

\icmlcorrespondingauthor{Pirzada Suhail}{psuhail@iitb.ac.in}

\icmlkeywords{Machine Learning, ICML}

\vskip 0.3in
]



\printAffiliationsAndNotice{} 

\begin{abstract}
In vision classification, generating inputs that elicit confident predictions is key to understanding model behavior and reliability, especially under adversarial or out-of-distribution (OOD) conditions. While traditional adversarial methods rely on perturbing existing inputs to fool a model, they are inherently input-dependent and often fail to ensure both high confidence and meaningful deviation from the training data. In this work, we extend network inversion techniques to generate Confidently Classified Counterfeits (CCCs), synthetic samples that are confidently classified by the model despite being significantly different from the training distribution and independent of any specific input. We alter inversion technique by replacing soft vector conditioning with one-hot class conditioning and introducing a Kullback-Leibler divergence loss between the one-hot label and the classifier’s output distribution. CCCs offer a model-centric perspective on confidence, revealing that models can assign high confidence to entirely synthetic, out-of-distribution inputs. This challenges the core assumption behind many OOD detection techniques based on thresholding prediction confidence, which assume that high-confidence outputs imply in-distribution data, and highlights the need for more robust uncertainty estimation in safety-critical applications.\end{abstract}

\section{Introduction}
\label{submission}
Neural networks have demonstrated exceptional performance across a wide range of applications, from image recognition to autonomous driving and medical diagnostics. Despite their success, these models are inherently uninterpretable with opaque decision-making processes. This lack of transparency is especially problematic in safety-critical settings, where trust, reliability, and interpretability are paramount. 

A central challenge in enhancing interpretability lies in understanding the relationship between model predictions and the input data. Commonly, high-confidence predictions are assumed to reflect the model’s exposure to in-distribution data—inputs similar to those seen during training. However, this assumption is not always valid. In reality, neural networks may confidently classify data that is significantly different from the training distribution, hereon referred to as Confidently Classified Counterfeits.

In this work, we introduce a novel approach for generating CCCs using network inversion. Network inversion typically works by reconstructing inputs that would lead to specific outputs, enabling insight into the model’s learned features. However, traditional inversion does not guarantee that the generated samples are confidently classified, which is crucial for understanding how the model behaves under confident predictions. Our method focuses on generating synthetic samples that are confidently classified by the model while being distinctly different from the training data, challenging the conventional notion that high-confidence predictions are always tied to in-distribution samples. Crucially, unlike adversarial examples that rely on perturbations of real inputs, CCCs are generated in a model-centric manner and are independent of any specific input.

To achieve this, we propose a conditioned generator that learns to produce diverse, plausible samples corresponding to specific class labels. We modify the conditioning mechanism from simple label vectors to one-hot vectors and matrices, which better capture the model's classification logic. To encourage diversity in the generated CCC samples, we apply a combination of loss functions, including cross-entropy, Kullback-Leibler (KL) divergence and cosine similarity to prevent redundancy in the generated set, resulting in a more varied set of counterfeits.

CCCs challenge the foundational assumption of many out-of-distribution detection techniques that rely on thresholding model confidence to distinguish in-distribution from OOD inputs. As such, this work contributes to improving the interpretability, robustness, and safety of neural networks—particularly in domains where overconfident errors could have serious consequences.

\section{Prior Work}

The idea of neural network inversion has attracted considerable attention as a means of gaining insight into the inner workings of neural networks. Inversion focuses on identifying input patterns that closely match a given output, allowing us to uncover the data processing functions embedded in the network’s parameters. These methods provide critical insights into how models represent and manipulate information, thereby exposing the latent structure within neural networks. Early work on inversion for multi-layer perceptrons, as demonstrated by \citet{KINDERMANN1990277} using the back-propagation algorithm, highlighted the utility of this technique in tasks like digit recognition. Despite the strong generalization abilities of multi-layer perceptrons—successfully classifying previously unseen digits—they often struggle to reject counterexamples, such as random input patterns.

Building on this, \citet{784232} introduced evolutionary inversion techniques for feed-forward networks, notable for their ability to identify multiple inversion points simultaneously. This approach provides a broader perspective on the input-output relationships within the network. \citet{SAAD200778} examined the lack of interpretability in artificial neural networks (ANNs) and proposed an inversion-based method for rule extraction, generating input patterns that correspond to specific output targets. This enables the creation of hyperplane-based rules that clarify the network’s decision-making process. In a different approach, \citet{Wong2017NeuralNI} addressed the inversion of deep networks by formulating the task of finding inputs that minimize certain output criteria as a constrained optimization problem, solved via the alternating direction method of multipliers. In adversarial settings, \citet{10.1145/3319535.3354261} investigated model inversion attacks, where an adversary infers training data by training a secondary network to reverse a model’s predictions, often using auxiliary datasets that are aligned with their prior knowledge.

Further contributions include Model Inversion Networks (MINs) introduced by \citet{NEURIPS2020_373e4c5d}, which learn an inverse mapping from scores to inputs for efficient handling of high-dimensional data. \citet{ansari2022autoinverseuncertaintyawareinversion} proposed a method that selects inverse solutions near reliable data points while incorporating predictive uncertainty to enhance inversion robustness and accuracy. To overcome inefficiencies of gradient-based inversion over complex loss landscapes, \citet{liu2022landscapelearningneuralnetwork} proposed learning a smoother surrogate landscape for improved convergence and stability. Meanwhile, \citet{suhail2024network} introduced a deterministic, constraint-based inversion method using SAT solvers on CNF encodings of the network, enabling the generation of diverse inversions at the expense of computational tractability.

In parallel, significant research has focused on generating adversarial samples—inputs intentionally crafted to elicit incorrect or high-confidence predictions from a model. A foundational contribution by \citet{goodfellow2015explainingharnessingadversarialexamples} demonstrated that small, imperceptible perturbations to inputs could cause models to misclassify with high confidence. Their linear explanation for adversarial examples revealed that neural networks are inherently vulnerable due to their high-dimensional linear structure. More recent work, such as \citet{grabinski2022robustmodelsoverconfident}, shows that robust models trained to withstand adversarial attacks tend to be less overconfident on out-of-distribution (OOD) data, emphasizing the link between robustness and calibrated confidence. Other approaches, like generating adversarial examples using adversarial networks (GANs), \citet{xiao2019generatingadversarialexamplesadversarial} aim to synthesize entire input distributions that mislead the model, not just perturbations of existing samples. These methods align closely with the broader goal of probing model decision boundaries but generally rely on input-dependent transformations.

Unlike most adversarial approaches that center around modifying real inputs, our work focuses on a model-centric strategy that generates Confidently Classified Counterfeits (CCCs)—samples that are confidently classified by the model without being tied to any real input. In this paper, we apply network inversion to generate CCCs that expose regions in the input space where the model is unjustifiably confident. Our method employs a conditioned generator trained to synthesize samples with high classifier confidence while being structurally distinct from the training data. To ensure diversity, we encode conditioning in a hidden form, apply heavy dropout during generation, and minimize cosine similarity between generated feature representations. Importantly, our method requires no access to training data and operates solely with access to the trained model. This allows us to generate diverse CCCs with varying confidence levels for any class, enabling analysis of the model’s behavior across its entire decision space. By challenging the assumption—commonly used in OOD detection methods—that high-confidence predictions indicate in-distribution data, our approach opens new pathways for auditing model reliability and exposing vulnerabilities in overconfident predictions.

\section{Network Inversion}

In this paper we use the Network Inversion approach proposed in \citep{suhail2024networkcnn} that uses a single carefully conditioned generator that learns data distributions in the input space of the trained classifier. This method relies solely on the input-output relationship of a fixed, trained classifier \( f_\theta: \mathcal{X} \rightarrow \Delta^{K-1} \), where \(\mathcal{X}\) is the input space and \( \Delta^{K-1} \) is the \((K-1)\)-D probability simplex over class labels.

\begin{figure*}[t]
\centering
\includegraphics[width=0.9\textwidth]{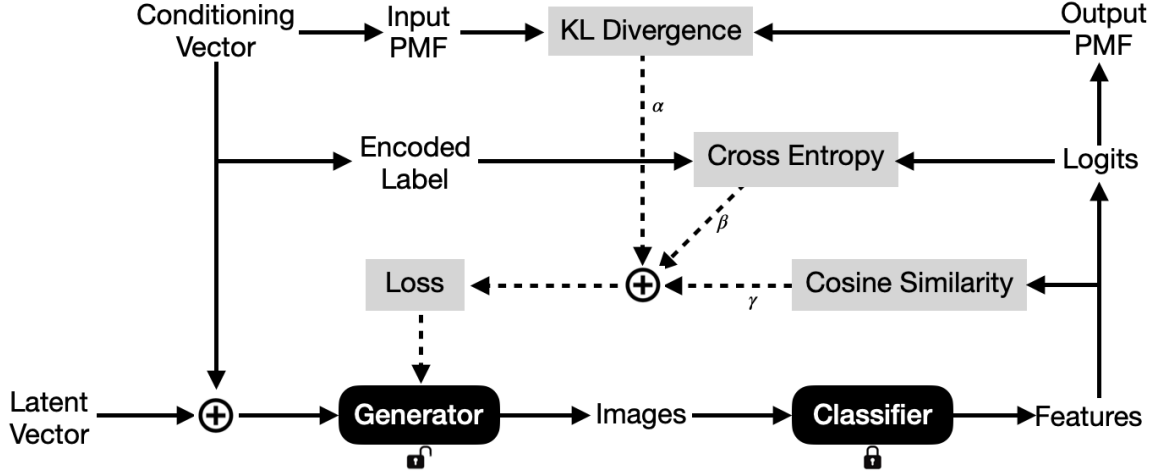} 
\caption{Proposed Approach to Network Inversion}
\label{1}
\end{figure*}

Formally, we train a conditional generator \( \mathcal{G}_\phi: \mathcal{Z} \times \mathbb{R}^K \rightarrow \mathcal{X} \), parameterized by \(\phi\), to invert the classifier’s behavior by optimizing it to minimize a composite loss

\[
\mathcal{L}_{\text{Inv}} = 
\alpha \cdot \mathcal{L}_{\text{KL}} +
\beta \cdot \mathcal{L}_{\text{CE}} +
\gamma \cdot \mathcal{L}_{\text{Cosine}}
\]

where, \( \mathcal{L}_{\text{KL}} \) is the KL Divergence loss, \( \mathcal{L}_{\text{CE}} \) is the Cross Entropy loss, and \( \mathcal{L}_{\text{Cosine}} \) is the Cosine Similarity loss. The hyperparameters \( \alpha, \beta, \gamma \) control the contribution of each individual loss term. They are defined as:
\[
\mathcal{L}_{\text{KL}} = \sum_{i} P(i) \log \frac{P(i)}{Q(i)}, \quad \mathcal{L}_{\text{CE}} = -\sum_{i} y_{i} \log(\hat{y}_{i}), \quad 
\]

\[
\mathcal{L}_{\text{Cosine}} = \frac{1}{N(N-1)} \sum_{i \neq j} \cos(\theta_{ij})
\]

where \( \mathcal{L}_{\text{KL}} \) represents the KL Divergence between the input distribution \( P \) and the output distribution \( Q \), \( y_{i} \) is the set encoded label, \( \hat{y}_{i} \) is the predicted label from the classifier, and \( \cos(\theta_{ij}) \) is the cosine similarity between the features of generated images \( i \) and \( j \) in a batch of \( N \).

\section{Confidently Classified Counterfeits}

Inversion and subsequent generation of counterfeits is performed on classifiers which include convolution and fully connected layers as appropriate to the classification task. We use standard non-linearity layers like Leaky-ReLU \citep{xu2015empiricalevaluationrectifiedactivations} and Dropout layers \citep{JMLR:v15:srivastava14a} in the classifier for regularisation purposes to discourage memorisation. The classification network is trained on a particular dataset and then held in evaluation mode for the purpose of inversion and generation of counterfeits.

The generator in our approach is conditioned on vectors and matrices to ensure that it learns diverse representations of the data distribution. Unlike simple label conditioning, the vector-matrix conditioning mechanism encodes the label information more intricately, allowing the generator to better capture the input space of the classifier.  The generator is initially conditioned using $N$-dimensional vectors for an $N$-class classification task. These vectors are derived from a normal distribution and are softmaxed to form a probability distribution. They implicitly encode the labels, promoting diversity in the generated images. Further, a Hot Conditioning Matrix of size $N \times N$ is used for deeper conditioning. In this matrix, all elements in a specific row or column are set to $1$, corresponding to the encoded label, while the rest are $0$.

In order to generate counterfeits, we modify the conditioning mechanism in inversion by shifting from soft vectors to one-hot vectors and matrices. In addition, we apply Kullback-Leibler Divergence (KLD) between the one-hot vectors and the output distribution of the classifier. This ensures that the generated samples are confidently classified by the model. With the classifier trained, the inversion is performed by training the generator to learn the data distribution for different classes in the input space of the classifier, as shown schematically in Figure \ref{1}.

\section{Results}

\begin{figure*}[h]
\centering
\includegraphics[width=1\textwidth]{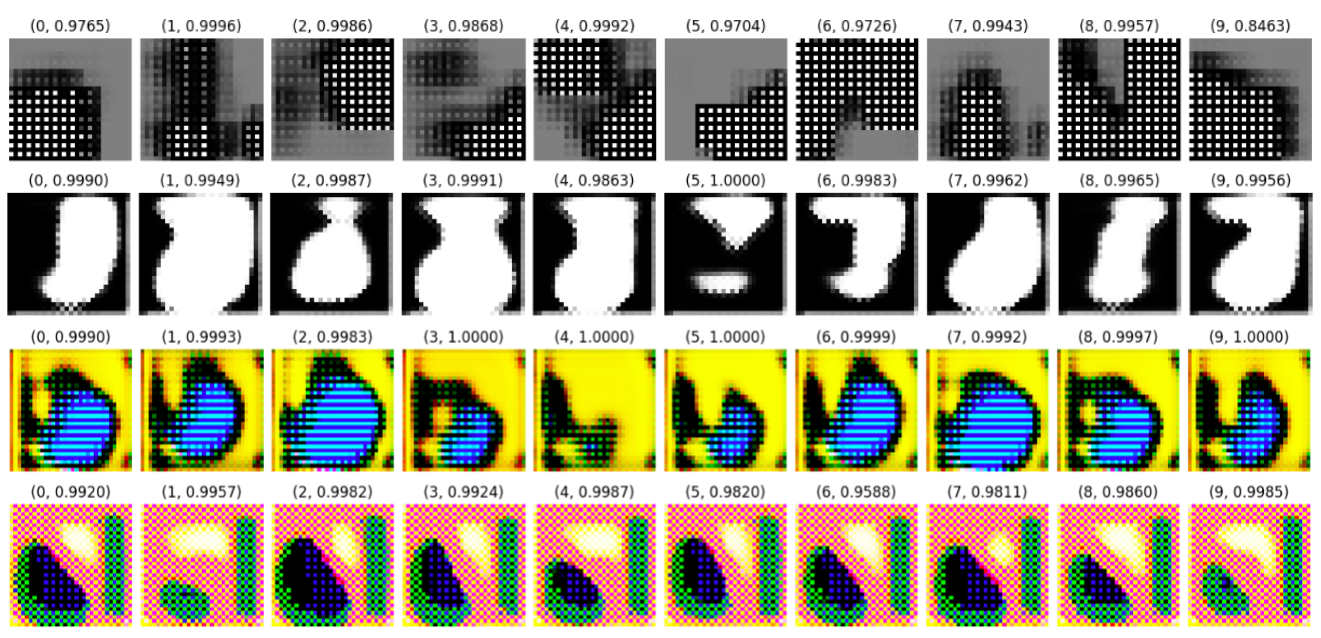}
\caption{Counterfeit images with the confidence values generated for all 10 classes in MNIST, FMNIST, SVHN \& CIFAR-10 respectively.}
\label{2}
\end{figure*}

The counterfeit generation experiments were carried out by modifying the inversion technique on the MNIST \citep{deng2012mnist}, FashionMNIST \citep{xiao2017fashionmnistnovelimagedataset}, SVHN and CIFAR-10 \citep{cf} datasets by training a generator to produce images that, when passed through a classifier, elicit the desired labels with very high confidence. The classifier is initially normally trained on a dataset and then held in evaluation for the purpose of inversion. The images generated by the conditioned generator corresponding to the latent and the conditioning vectors are then passed through the classifier to compute the loss between the classifier outputs and the conditioning mechanism.

The classifier is a simple multi-layer convolutional neural network consisting of convolutional layers, dropout layers, batch normalization, and leaky-relu activation followed by fully connected layers and softmax for classification. While the generator is based on Vector-Matrix Conditioning in which the class labels are encoded into random softmaxed vectors concatenated with the latent vector followed by multiple layers of transposed convolutions, batch normalization \citep{pmlr-v37-ioffe15} and dropout layers \citep{JMLR:v15:srivastava14a}. Once the vectors are upsampled to \(NXN\) spatial dimensions for an N class classification task they are concatenated with a conditioning matrix for subsequent generation upto the required image size of 28X28 or 32X32 as required.

The distribution of Confidently Classified Counterfeits generated along with their confidence values are visualized in Figure \ref{2} for all 10 classes of MNIST, FashionMNIST, SVHN, and CIFAR-10, respectively. It can be observed that the model is near perfectly confident in its classification of these samples, as reflected in the softmax scores. Despite being confidently classified, these images are distinctly different from anything the model was trained on, yet they fall within the input space of different labels, highlighting their unsuitability for safety-critical tasks. 

In applications, where models are expected to demonstrate high confidence only on in-distribution data, the ability to confidently classify out-of-distribution inputs presents a critical vulnerability. These Confidently Classified Counterfeits expose the model’s susceptibility to miss-classifying unfamiliar inputs as high-confidence predictions, raising concerns about the model’s reliability in safety-critical environments, where erroneous predictions could have severe consequences. This reveals the need for more robust methods to address out-of-distribution data to ensure the model’s trustworthiness in real-world applications.

\section{Conclusions}

In conclusion, the generation of Confidently Classified Counterfeits (CCCs) reveals critical vulnerabilities in neural networks, especially in safety-critical settings where reliability is essential. By extending network inversion techniques, we demonstrate that a model can assign high confidence to synthetic inputs that are structurally and statistically distant from the training distribution, directly challenging the common assumption that high-confidence predictions imply in-distribution data. Unlike traditional adversarial or inversion approaches, our method is entirely model-centric as it requires no access to training data and can generate diverse CCC samples with near-perfect confidence scores. This work highlights the need to reassess confidence-based assumptions in model behavior and the limitations of existing out-of-distribution detection methods that rely on confidence thresholds.


\bibliography{example_paper}
\bibliographystyle{icml2025}

\end{document}